\documentclass[subscriptcorrection,upint,varvw,barcolor=Goldenrod3,mathalfa=cal=euler,balance,hyphenate,pdf-a,nofoot,nolists]{asmejour} %

\hypersetup{
	colorlinks=true,
	linkcolor=blue,
	citecolor=blue,
	urlcolor=blue,
}

\usepackage{float}
\usepackage{amsmath}
\usepackage{booktabs}
\usepackage{multirow}
\usepackage{graphicx}
\usepackage{colortbl}
\usepackage{gensymb}
\usepackage{makecell} %
\usepackage{pifont} %
\newcommand{\cmark}{\ding{51}} %
\newcommand{\xmark}{\ding{55}} %

\usepackage{tabularx}

\definecolor{customblue}{RGB}{149,203,251}
\usepackage{mdframed}
\newmdenv[
  topline=false,
  bottomline=false,
  rightline=false,
  linewidth=2pt,
  linecolor=customblue,
  backgroundcolor=gray!20,
  leftmargin=10pt,
  rightmargin=10pt,
  innertopmargin=10pt,
  innerbottommargin=10pt,
  font=\small
]{customquote}

\newcommand{\acr}{DesignQA}

\begin{document}

\JourName{}%

\SetAuthorBlock{Anna C. Doris\CorrespondingAuthor}{%
Massachusetts Institute of Technology, \\
77 Massachusetts Avenue,\\
Cambridge, MA 02139, USA\\
email: adoris@mit.edu
}

\SetAuthorBlock{Daniele Grandi}{%
Autodesk Research, \\
The Landmark @ One Market, Ste. 400,\\
San Francisco, CA 94105, USA
}

\SetAuthorBlock{Ryan Tomich}{%
MIT Motorsports, \\
77 Massachusetts Avenue,\\
Cambridge, MA 02139, USA
}

\SetAuthorBlock{Md Ferdous Alam}{%
Massachusetts Institute of Technology, \\
77 Massachusetts Avenue,\\
Cambridge, MA 02139, USA
}

\SetAuthorBlock{Mohammadmehdi Ataei}{%
Autodesk Research, \\
661 University Avenue,\\
Toronto, Ontario M5G 1M1, Canada
}

\SetAuthorBlock{Hyunmin Cheong}{%
Autodesk Research, \\
661 University Avenue,\\
Toronto, Ontario M5G 1M1, Canada
}

\SetAuthorBlock{Faez Ahmed}{%
Massachusetts Institute of Technology, \\
77 Massachusetts Avenue,\\
Cambridge, MA 02139, USA
}

\title{\NoCaseChange{DesignQA}: A Multimodal Benchmark for Evaluating Large Language Models’ Understanding of Engineering Documentation} %

\begin{abstract}
This research introduces \acr{}, a novel benchmark aimed at evaluating the proficiency of multimodal large language models (MLLMs) in comprehending and applying engineering requirements in technical documentation. Developed with a focus on real-world engineering challenges, \acr{} uniquely combines multimodal data—including textual design requirements, CAD images, and engineering drawings—derived from the Formula SAE student competition. Different from many existing MLLM benchmarks, \acr{} contains document-grounded visual questions where the input image and input document come from different sources. The benchmark features automatic evaluation metrics and is divided into segments—Rule Comprehension, Rule Compliance, and Rule Extraction—based on tasks that engineers perform when designing according to requirements. We evaluate state-of-the-art models (at the time of writing) like GPT-4o, GPT-4, Claude-Opus, Gemini-1.0, and LLaVA-1.5 against the benchmark, and our study uncovers the existing gaps in MLLMs' abilities to interpret complex engineering documentation. The MLLMs tested, while promising, struggle to reliably retrieve relevant rules from the Formula SAE documentation, face challenges in recognizing technical components in CAD images, and encounter difficulty in analyzing engineering drawings. These findings underscore  the need for multimodal models that can better handle the multifaceted questions characteristic of design according to technical documentation. This benchmark sets a foundation for future advancements in AI-supported engineering design processes. \acr{} is publicly available at: \href{https://github.com/anniedoris/design\_qa/}{https://github.com/anniedoris/design\_qa/}.
\end{abstract}

\date{}
\maketitle

\section{INTRODUCTION}
Large language models (LLMs), such as ChatGPT~\cite{gpt4v}, are chat-bots that can engage in conversations based on user queries. Trained on data from much of the internet, LLMs are based on the Transformer architecture \cite{vaswani2017attention} and have learned to predict the next words based on an  input sequence of text \cite{bubeck2023sparks}. ChatGPT is the fastest adopted technology in history, with more than 100 million users two months after its release \cite{barbhuiya2023introduction}. LLMs have garnered significant attention for their conversational abilities, and research studies have examined and quantified their abilities to answer questions on a range of topics, from medicine~\cite{thirunavukarasu2023large, clusmann2023future} to education~\cite{kasneci2023chatgpt} to engineering~\cite{makatura2023can, picard2023concept}. 

With the emergence of LLMs as conversational assistants, an important question is how they can help humans answer questions about engineering design problems. A key goal of design automation has been to have an AI helper that can make it easier and faster for human designers to create better products. Although Generative AI has made significant strides, this goal has been difficult to attain, since engineering design tasks necessitate synthesis of multimodal information across multiple sources. One such task, critical to engineering design, is designing products based on technical requirements, which list rules that consist of a metric and a value (e.g. maximum tire width can be no greater than five inches)~\cite{ulrich2016product}. Matching the complexity of many real-world designs, technical specifications can be lengthy and extremely detailed and often reference critical safety or regulatory specifications. Designing according to requirements necessitates that engineers or designers can interpret and synthesize multimodal data across sources (e.g. the requirements document, CAD, engineering drawings, documentation, standards, etc.).

\begin{figure*}[t]
\begin{center}
\setlength{\unitlength}{0.012500in}%
\includegraphics[width=\linewidth]{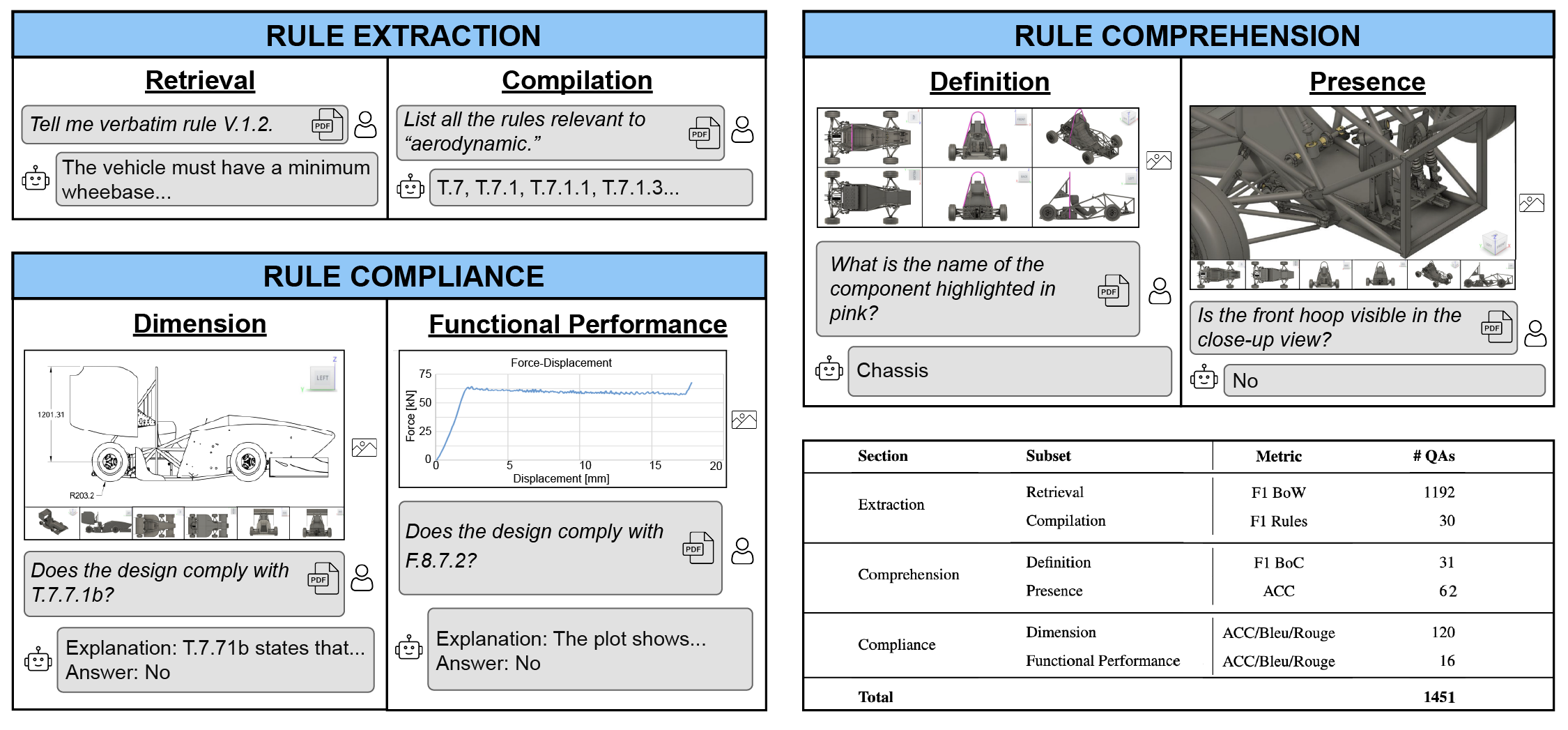}
\end{center}
\caption{Overview of the three different segments (Rule Extraction, Rule Comprehension, and Rule Compliance) and six subsets (Retrieval, Compilation, Definition, Presence, Dimension, and Functional Performance) in \acr{}. Prompts and images shown above are condensed versions of the actual prompts and images used. The bottom right table shows the metrics and the number of questions for each subset of the benchmark.}
\label{fig:techspec_overview} 
\end{figure*}

Recently, models with multimodal capabilities~\cite{zeng2023matters, wang2023t, liu2024visual}, lengthy documents (long-text) processing capabilities~\cite{shaham2023zeroscrolls, xiong2023effective}, and both multimodal \textit{and} long-text capabilities~\cite{gpt4o, gpt4v} have been developed. These advances bring us closer to the reality of a multimodal AI assistant that could help automate engineering design according to requirement documents. As new MLLMs emerge, evaluating their capacity to fulfill these essential design-requirement-related tasks becomes imperative. This begs the question: How good are contemporary MLLMs at engineering design according to requirements? How can we measure tangible improvements in the efficacy of MLLMs at these tasks? Thus, we propose a novel benchmark aimed at assessing the proficiency of MLLMs in interpreting and adhering to the complex and multimodal demands of technical requirements in the design process.

We present \acr{} (Figure \ref{fig:techspec_overview}), the first zero-shot benchmark for engineering design technical requirements question-answering. The benchmark consists of 1451 questions and is based on the Formula SAE 2024 Rules and data (CAD, documentation, etc.) provided by the MIT Motorsports team. By developing this benchmark in conjunction with the MIT Motorsports team, we prioritized the generation of a dataset that is characteristic of real-world design requirement challenges. \acr{} also contains document-grounded reference-dependent visual question-answers (VQAs), one of a handful of benchmarks that tests models' abilities to answer questions that require analysis across long-text documents \textit{and} images. Notably, our benchmark assesses a model's ability to synthesize information across an image and text from different sources, where the image was not seen by the model during its original training (pre-training).

In addition to developing the dataset, we used \acr{} to benchmark several state-of-the-art (at the time of writing\footnote{All references to ``time of writing'' refer to the period from April - August 2024.}) MLLMs, GPT-4o \cite{gpt4o}, GPT-4 \cite{gpt4v}, Gemini-1.0 \footnote{https://ai.google.dev/gemini-api/docs/models/gemini?authuser=1}, Claude-Opus \footnote{https://www.anthropic.com/news/claude-3-family}, and LLaVA-1.5 \cite{liu2023llava}, providing the FSAE rules to the models either via the context window or via a simple retrieval method. While GPT-4o is generally the best performer -- of the models tested -- on DesignQA, our findings highlight the need for MLLMs that can better perform the multifacted tasks of engineering design: for example, a model that can better parse long text \textit{and} analyze images \textit{and} apply technical knowledge, simultaneously. Based on observations of the performances of these models on the benchmark, we provide suggestions about how models might be modified for improved results on \acr{} and design requirement questions generally.

In summary, our contributions are:

\begin{enumerate}
    \item \textbf{A Novel, Multifaceted Benchmark for MLLMs}: We introduce \acr{}, a benchmark that tests MLLMs' understanding of design according to an engineering requirement document. \acr{} is unique in its need for models to analyze and integrate information from both visual and long-text inputs, emphasizing the complexity and multimodal nature of real-world engineering tasks.
    \item \textbf{A Granular and Automatic Evaluation Framework}: We create \acr{} to be thorough and easy to use. The benchmark is divided into three segments - rule extraction, comprehension, and compliance - enabling a fine-grained investigation into a model's strengths and weaknesses and enriching our understanding of AI in technical domains. Each subset of \acr{} has an automatic evaluation metric, permitting the quick evaluation of future MLLMs.
    \item \textbf{High Quality, Real-World Question-Answer Pairs}: We develop a high-quality benchmark based on real-world data and problems. The question-answers in \acr{} are based on the FSAE competition rules and data provided by the MIT Motorsports team. Questions are designed and reviewed by members of the MIT Motorsports team, industry professionals, and engineering researchers.
    \item \textbf{Evaluation of Contemporary MLLMs}: We evaluate GPT-4o, GPT-4, Gemini-1.0, Claude-Opus, and LLaVA-1.5 against \acr{}, unveiling the current limitations of AI and retrieval methods in handling multimodal data and processing engineering requirements. Some difficulties that the tested models face include reliably extracting rules from the documentation, recognizing technical components in CAD images, and analyzing engineering drawings.

\end{enumerate}

\section{RELATED WORK}

In this section, we first provide an overview of existing work on AI for engineering design, showcasing that MLLMs have new potential to assist humans with design and design requirement problems. We then explore existing benchmarks for LLMs and MLLMs, highlighting the lack of benchmarks about engineering design and design requirements. We then categorize existing benchmarks by reference type. This sets the context for our contribution, which addresses the need for real-world, document-grounded benchmarks that bridge textual and visual information comprehensively.

\begin{table*}[t]
\caption{Overview of select LLM and MLLM benchmarks, their domains, and reference-dependence. Our benchmark is unique in its focus on design requirements, and in that it contains multi-source document-grounded VQAs.}
\label{tab:related_work}
\begin{center}
\setlength\tabcolsep{5pt} %
\renewcommand{\arraystretch}{1.2} %
\begin{tabular}{@{}lcccccc@{}}
\toprule
 &  & \multicolumn{4}{c}{Reference Type} &  \\ \cline{3-6}
Benchmark & Domain & Self-contained & Open-domain & \makecell{Doc-Grounded \\ Single Source*} & \makecell{Doc-Grounded \\ Multi Source**} & VQA \\
\midrule
McTest~\cite{richardson2013mctest} & Narrative Children's Stories & \cmark & - & - & - & \xmark \\
SQUAD~\cite{rajpurkar2016squad} & \makecell{Wikipedia} & - & \cmark & - & - & \ding{55} \\
WikiQA~\cite{yang2015wikiqa} & \makecell{Wikipedia} & - & \cmark & - & - & \ding{55} \\
QASPER~\cite{dasigi2021dataset}&  \makecell{NLP Papers} & - & - & \cmark & - & \ding{55} \\
ZeroScrolls~\cite{shaham2023zeroscrolls} & \makecell{Mixed: Wiki, Gov, etc.} & - & - & \cmark & - & \ding{55} \\
MME~\cite{fu2023mme} & \makecell{COCO} & \cmark & - & - & - & \cmark \\
MM-Bench ~\cite{liu2023mmbench} & \makecell{Mixed: COCO, Llava, etc.} & \cmark & - & - & - & \cmark \\
MMMU~\cite{yue2023mmmu} & \makecell{College materials} & - & \cmark & - & - & \cmark \\
ScienceQA~\cite{lu2022learn} & \makecell{Open-source science materials} & - & \cmark & - & - & \cmark \\
InfoSeek~\cite{chen2023can} & \makecell{Wikipedia} & - & - & \cmark & - & \cmark \\
\textbf{\acr{} (Ours)}& \makecell{\textbf{FSAE Rules Doc \& Data}} & - & - & - & \cmark & \cmark \\
\bottomrule
\end{tabular}
\end{center}
\footnotesize{*Single source: image in question contained within the document; **Multi source: image in question not contained within the document}
\end{table*}

\subsection{AI for Engineering Design}
Much of the prior work on AI for design has focused on single modalities~\cite{song2024multi}, such as images or text. For text, several studies have investigated natural language processing (NLP) for technical engineering text. For example, \cite{dima2021adapting} and \cite{brundage2021technical} describe a technical language processing framework for circumnavigating the limitations of traditional NLP on unstructured engineering data. Expanding beyond text, \cite{jiang2022deep} creates a deep learning architecture for technical document classification that factors in images as well as text. Despite significant advancements, many NLP and deep learning methods are specialized to a single domain and don't generalize well to other problems within engineering design. 

LLMs offer more generalizable solutions to various problems within engineering design. \cite{zhu2023generative} demonstrates the potential for LLMs (GPT-2 and GPT-3) to automate early-stage design concept generation. \cite{makatura2023can} explores how LLMs can assist engineers across an array of design and manufacturing tasks. While LLMs are useful for select engineering design tasks, many engineering tasks are highly multimodal (involving images, CAD, graphs, etc.). Therefore, recent advancements in MLLMs hold untapped potential for the automation of engineering design tasks. \cite{picard2023concept} investigates the potential of GPT-4 to automate engineering design tasks involving images, creating a dataset of over 1000 zero-shot queries. However, this dataset does not focus on engineering documentation. While a plethora of AI models like Google's Gemini, Meta's Llama family, and Anthropic's Claude models have emerged recently, their effectiveness is almost exclusively evaluated on non-engineering benchmarks. Critical for characterizing the abilities of MLLMs for engineering design tasks are benchmarks that can rigorously quantify their performances, which serves as inspiration for \acr{}.

\subsection{LLM and MLLM Benchmarks}
In this section, we explore existing benchmarks for LLMs and MLLMs based on domain and reference type. See Table \ref{tab:related_work} for a concise overview.

\subsubsection{Benchmarks for Engineering Design and Design Requirements}
Very few benchmarks exist for engineering design problems or design requirement-related tasks. Despite the plethora of complex, multimodal QAs that could be generated from technically rich design requirement documents, very few datasets or benchmarks have been developed for this domain. PURE (PUblic REquirements)~\cite{ferrari2017pure} is a dataset composed of 79 requirements documents scraped from the web. However, the dataset does not provide QA pairs and thus cannot be easily used for benchmarking. \acr{} harnesses the FSAE competition rules and MIT Motorsport design data to develop a benchmark of QAs pertaining to real-world design requirements.

\subsubsection{LLM Reference-Dependent Benchmarks}
Text-based QA benchmarks that require a model to parse additional references to answer the posed question can be called ``reference-dependent'' benchmarks. Reference-dependent benchmarks differ from many classic reading comprehension benchmarks, like MCTest~\cite{richardson2013mctest}, which are ``self-contained'' and can be answered by short-text (approximately paragraph length) chunks accompanying the question. Since reference-dependent benchmarks require a model to locate the relevant information -- usually across one or multiple long texts -- and then apply that information to the posed question, they tend to be more complex questions. Following the distinction made by Dasigi \textit{et al.}~\cite{dasigi2021dataset}, reference-dependent benchmarks can further be divided into ``open-domain'' benchmarks and ``document-grounded'' benchmarks. An example of an open-domain question, taken from SQUAD, is: ``Where do water droplets collide with ice crystals to form precipitation?''~\cite{rajpurkar2016squad} Open-domain benchmarks, such as SQUAD~\cite{rajpurkar2016squad} and WikiQA~\cite{yang2015wikiqa}, test a model's ability to answer general-knowledge, factoid-type questions, the answer for which is usually contained in multiple sources in the model's pre-trained corpus.

In contrast, document-grounded benchmarks, like QASPER~\cite{dasigi2021dataset} and ZeroScrolls~\cite{shaham2023zeroscrolls}, test a model's ability to answer questions based on information provided in a specific long-text document. An example of a document-grounded question, taken from QASPER, is: ``[In reference to a specific NLP paper] Which neural architecture do they use as a base for their attention conflict mechanisms?''~\cite{dasigi2021dataset} As noted by Dasigi \textit{et al.}, document-grounded QAs tend to be more complex than open-domain QAs since they are anchored in user context and the answers are not widely available facts~\cite{dasigi2021dataset}. \acr{}, grounded in the FSAE rule document, fits within this document-grounded category. As a result, the questions posed in our benchmark are complex and rooted in user needs rather than common sense. Document-grounded questions are very characteristic of engineering design problems, as various types of documents -- standards, manuals, documentation, etc. -- often contain specific information that cannot be easily found on the internet. 

\subsubsection{MLLM Benchmarks}
Multimodal datasets typically test an MLLM's capacity to analyze a non-text element with respect to question text. At the time of writing, most MLLMs can only accept images as non-text inputs, so multimodal QA benchmarks tend to consist of a visual (image) coupled with a question/answer pair. Visual question-answers (VQAs) can be categorized in the same way as text-based QA benchmarks. The vast majority of VQA benchmarks are self-contained. An example of a self-contained VQA question, taken from MME, is: ``[Pertaining to a photo showing doubles tennis partners] Are there two people in this image?'' These are questions for which no  context (other than the image) is needed to answer the question. The MME~\cite{fu2023mme} and MMBench~\cite{liu2023mmbench} benchmarks -- highlighted by Chang and Wang et al.'s review paper~\cite{chang2023survey} -- are both self-contained VQAs. These benchmarks largely focus on basic tasks -- primarily reasoning and perception -- which are usually presented in a multiple choice format for ease of evaluation~\cite{fu2023mme}.

More challenging and less prevalent than self-contained VQAs, reference-dependent VQAs test a model's ability to synthesize image analysis \textit{with} additional knowledge, either from the open domain or from specific documents. MMMU~\cite{yue2023mmmu}, which consists of multiple choice questions gathered from college materials, can be considered an open-domain VQA, as can ScienceQA~\cite{lu2022learn}, which contains elementary to high school-level science multiple choice questions. The benchmark additionally encourages complete ``train-of-thought'' reasoning by providing a ``lecture'' -- multiple sentences of general knowledge pertaining to the question -- and ``explanation'' -- reasoning for selecting a particular answer -- for each VQA. InfoSeek~\cite{chen2023can} is the first document-grounded VQA, composed of questions about images in specific Wikipedia articles that can only be answered by consulting the corresponding article's text. 

However, there is still a significant need for document-grounded VQAs that are more characteristic of real-world tasks. InfoSeek has a direct match between visual and the document (i.e. the visual is contained within the document), while most visual questions asked by users would not fit this direct look-up framework (e.g. provided with an image of a broken machine and a manual for the machine, the exact image will not be contained within the manual). InfoSeek's images and documents -- which both come from Wikipedia -- have also likely been seen by the model during pre-training; for many of the questions asked by users, either the document or the visual would not have been seen during pre-training. \acr{} is constructed to fill these gaps and is more representative of these real-world scenarios.

\section{\acr{} BENCHMARK}

\subsection{The Dataset}
We created DesignQA to be characteristic of real-world engineering problems, in three ways. Firstly, the benchmark is based on a real rulebook, the 140-page Formula SAE Rule document. While designed for a student competition, the Formula SAE Rule document guides young engineers in designing, building, testing, and racing a fully functional gas-powered or electric vehicle. Student engineers repeatedly consult the rule document while developing their vehicle prior to competition.  Secondly, the benchmark was created with real CAD models and test data provided by the MIT Motorsports team, the equivalent of which from engineering companies is not readily available. Thirdly, the QAs were created by a member of the MIT Motorsports team, a member of our team from industry (Autodesk), or a member of our team from academic research. All manually generated QAs -- except those that are derivatives of other questions or Rule Compliance explanations -- were reviewed by the two parties that didn't write the question. The importance of these second and third points cannot be overstated, as many LLM and MLLM datasets rely on crowdsourcing or synthetic data~\cite{wang2022self}, which is often generated by other LLMs. While much more straightforward to generate, crowdsourcing and synthetic data -- especially in highly technical domains -- often comes at the cost of QA quality.

The Formula SAE Rule document is similar to many other engineering design rule documents and standards. In fact, the 2024 Formula 1 Technical Regulations document~\cite{f1} -- which governs real Formula 1 races -- is 178-pages long and has very similar content sections as the student SAE version. As such, we believe that DesignQA and its subject matter well capture engineering-design-type questions. The Formula SAE Rule document is organized in sections (e.g. V-``Vehicle Requirements,'' F-``Chassis and Structural,'' etc.) with numbered rules (e.g. V.1, V.2, etc.) that are grouped logically. This is a common format for other technical documentation, such as NASA technical memorandums~\cite{kovich2023human}, which also follow similar enumerated structures. Like the Formula SAE rule document, many of the rules specified in these other documents refer to technical terms defined elsewhere within the document and many of the rules pertain to dimensional constraints or functional performance.

DesignQA consists of 1451 question-answers (Figure 1) and is divided into three segments -- Rule Extraction, Rule Comprehension, and Rule Compliance -- which build on each other and are designed to test MLLMs on the process of designing according to technical documentation. Given a rule document or standard, engineers repeatedly consult the document throughout the design process to check for design \textit{compliance}. Likewise, the Rule Compliance segment of the benchmark assesses an MLLM's capacity to check engineering drawings and test data for rule adherence. The other two segments of the benchmark test models on skills that are prerequisites for being able to evaluate compliance. For example, in order for engineers to check for compliance, they must first \textit{comprehend} all of the terms and definitions presented within the rule document. The Rule Comprehension segment of the benchmark evaluates an MLLM's ability to recognize in visual designs technical jargon from the rule document. Similarly, in order to check for rule compliance, engineers must be able to \textit{extract} -- from the rule document -- those rules that are relevant to their design task. In a similar way, the Rule Extraction questions test an MLLM's ability to identify relevant information in a lengthy rule document; without being able to perform these relatively simple identification tasks, models will be unable to answer any meaningful questions about a rule-in-question.

To provide more context to the model, each of the QAs begins with the following preamble:

\begin{customquote}
We are a student engineering team designing a vehicle for the FSAE competition. Attached is the FSAE rules document.
\end{customquote}

The rule document was provided to the model when asking each question. Each of the three benchmark segments is further divided into two subsets, or specific tasks. Details about each of the benchmark segments and subsets are presented in the following section.

\begin{figure*}[t]
\begin{center}
\setlength{\unitlength}{0.012500in}%
\includegraphics[width=\linewidth]{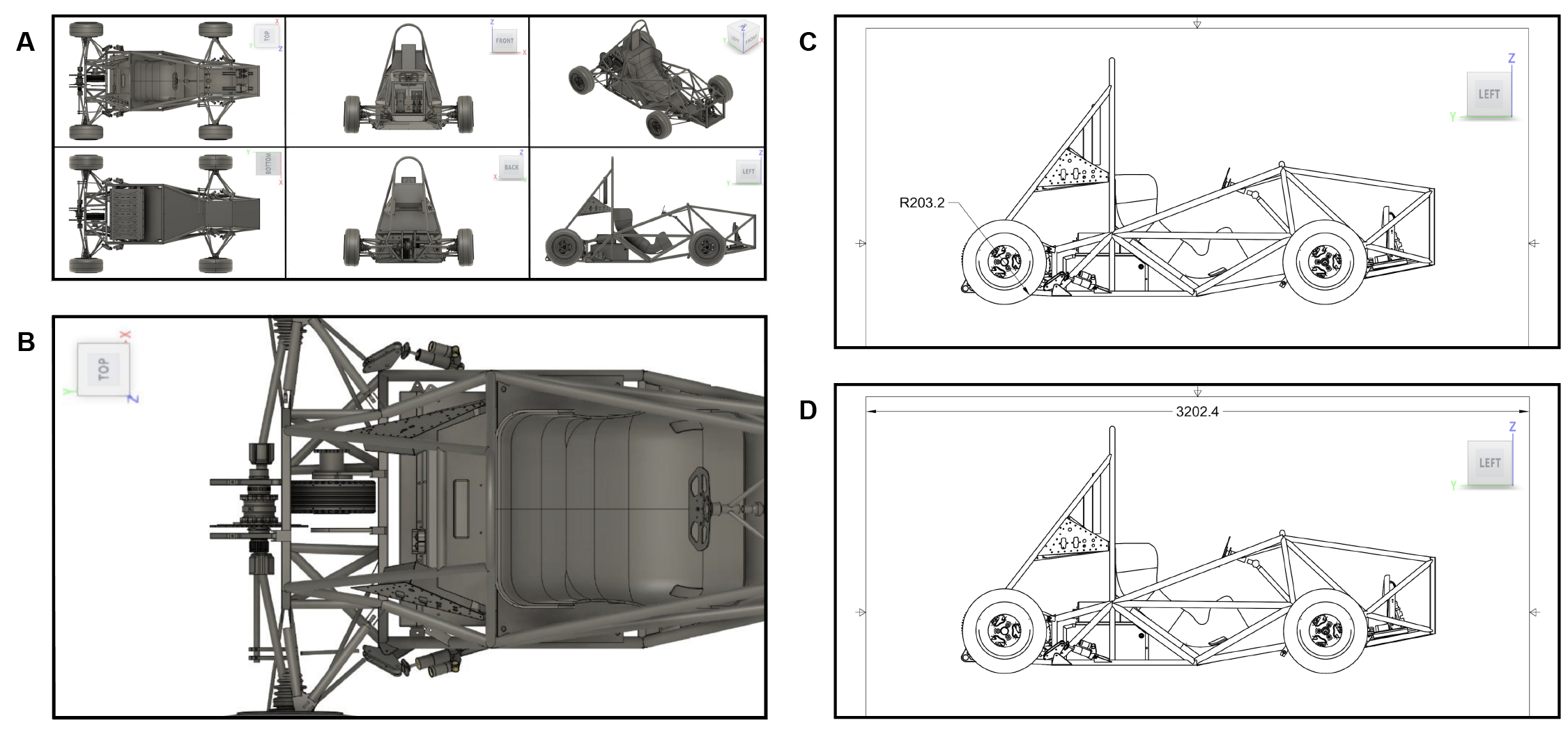}
\end{center}
\caption{Representing 3D CAD models in 2D images. A) Multi-view CAD image. B) Close-up CAD image. C-D) Engineering drawing images. C) uses the direct dimensioning method and D) uses the scale bar dimensioning method.}
\label{fig:cad_rep} 
\end{figure*}

\subsubsection{Rule Extraction}
A key - albeit usually simple - task for engineers is to locate a specific rule in a requirement document: answering a question about a rule is predicated on the ability to locate and extract the relevant rule. The Rule Extraction segment of the benchmark tests a model's ability to extract rules from the 140-page FSAE rule document. This benchmark segment is further divided into two subsets: Retrieval QAs and Compilation QAs.

\textbf{Retrieval} questions test a model's ability to extract specific information from a lengthy document. Given the large number of pages in the original rule document, retrieving the text of the rules word-for-word is a non-trivial task. While this retrieval task might become obsolete in the future as better models continue to be developed, it is critical for models to retrieve accurate information, as it is a necessary precursor for the other types of questions in this benchmark.

We programmatically create the Retrieval QAs by first extracting all the text from the PDF document, excluding the headers, footers, and page numbers. The rule document is well organized into numbered sections and subsections (which may or may not have titles) in the format `AA.\#.\#.\#'. By using a combination of manually created scripts and regex patterns to identify the individual rules, we tabulate the set of rules and label the rule number, the rule title, and the rule text. Finally, we drop the rules that do not contain any text (while keeping the child rules) as well as sets of rules that pertain to other aspects of the race (e.g., Administrative Regulations, Document Requirements) and not to the design specifications that the vehicle must meet.

From the tabulated set of rules, we can then programmatically formulate the set of Retrieval QAs. To the preamble described in the previous section, we append the following:

\begin{customquote}
What does rule $\{rule\_number\}$ state exactly? Answer with only the text of the rule and no other words.
\end{customquote}

Where $\{rule\_number\}$ is replaced by each of the selected rule numbers. This process results in 1192 Retrieval QAs.

\textbf{Compilation} questions assess a model's ability to look for information spanning a long document. A common task that designers might perform when interacting with a design requirement document is the compilation of all rules relevant to a specific subject, such as the `suspension' or `critical fasteners'. To create this set of QAs, we begin with a manually curated set of 30 common terms present in the rule document (nine of which include their synonyms, acronym, or plurals). This results in 30 questions with the following format:

\begin{customquote}
Please list all rules relevant to $\{term\}$. Answer with only the rule numbers (i.e.: AA.1.1.1) separated by commas and no other words.
The rules relevant to $\{term\}$ are:
\end{customquote}

Where $\{term\}$ is replaced by each of 30 common terms. 
To create the ground truth answers, we first compile the list of rules that include the term with a simple search through the tabulated rules, described in the previous section. We also include the children of each of the rules found, as well as any other rules that might be mentioned in both the parent and child rules.

\subsubsection{CAD Representation}

\label{sec:cad_rep}
The Rule Comprehension and Rule Compliance segments of our benchmark ask questions about 3D CAD models of the designed vehicle. We develop QAs around four different 3D CAD models provided by MIT Motorsports: the vehicle, the vehicle plus the aerodynamic package, the rear wheel package, and the powertrain. Before describing the details of these QAs, we devote this section to detailing how we provide 3D CAD model information to MLLMs. Since MLLMs cannot accept typical 3D CAD model formats (e.g. .stl, .step, etc.) at this time, we convert the CAD that we would like to show the model into 2D image forms, preserving as much 3D spatial information as possible. We represent 3D CAD models in 2D using three different kinds of images: 1) multi-view CAD images, 2) close-up CAD images, and 3) engineering drawing images (Figure \ref{fig:cad_rep}). 

\textbf{Multi-view CAD images} (Figure \ref{fig:cad_rep}A) show six views of the CAD model: top, bottom, front, back, left, and isometric. Since the model cannot rotate a 3D CAD model in a CAD software GUI, these six views capture information about how the different views fit together to comprise the 3D model. Each view has a corresponding coordinate frame, so that it is clear to the viewer how each of the six views is related to the others.

\textbf{Close-up CAD images} (Figure \ref{fig:cad_rep}B) show zoomed-in views of our CAD. The purpose of these images is to show finer detail in specific regions of the model. The close-up CAD images show a single view of the model with an orientation (and coordinate frame) matching one of the views in the corresponding multi-view CAD image.

\textbf{Engineering drawing images} (Figure \ref{fig:cad_rep}C\&D) display dimensional information about the 3D model. These images are created using engineering drawing software, so that the dimensions shown are highly accurate. They show a single view of the model with an orientation (and coordinate frame) matching one of the views in the corresponding multi-view CAD image. We used two different dimensioning systems to indicate the dimensions on these images. The first method was direct-dimensioning (as in Figure \ref{fig:cad_rep}C), where dimensions relevant to a particular rule are explicitly indicated on the drawing. The second method was scale-bar-dimensioning (as in Figure \ref{fig:cad_rep}D), where a scale bar is provided and from which a model could infer necessary dimensions. We used a mixture of these two dimensioning methods in our QAs, as we were interested in what effect the dimensioning method would have on model performance.

In the following sections, we describe how these three image types are employed in our QAs. Often, two of these image types are appended together to form an image that conveys more information. While we represent 3D CAD models using various 2D image types, this 3D model representation should be updated as MLLMs become more sophisticated and are able to parse inherent 3D model file formats.

\subsubsection{Rule Comprehension}
\label{sec:comprehension}
In order to understand how the rules relate to a design, engineers must first understand the terms presented in the rules and the names of the different components in the design. The Rule Comprehension segment of the benchmark evaluates a model's ability to refer to elements of a 3D model according to the definitions and terminology presented in the rule document. This part of the benchmark is further divided into two subsets: Definition QAs and Presence QAs.

\textbf{Definition} questions test a model's ability to identify the name of a highlighted component in a CAD model. From a list of 31 components, we created a multi-view CAD image (Figure \ref{fig:cad_rep}A) where the component-to-be-identified is highlighted in pink. Component synonyms were also collected (e.g. frame and chassis) for scoring purposes. Sometimes, it was necessary to hide some components in the CAD model so that the highlighted component could be better visualized. If components were hidden, it was noted in the prompt. Appended to the preamble is the following prompt, which resulted in the generation of 31 VQA pairs:

\begin{customquote}
Also attached is an image showing six CAD views of our vehicle design. What is the name of the component(s) highlighted in pink? \{[\textit{If components hidden}] Some parts of the design have been hidden so that the highlighted component(s) can better be visualized.\} Answer just with the name of the highlighted component(s) and nothing else.
\end{customquote}

\textbf{Presence} questions assess a model's ability to understand whether a particular component is present or not in a close-up CAD image. As such, these QAs are an easier variant of the Definition QAs, since they ask the model to provide a yes/no answer rather than the name of a component. Using the same list of 31 components from the Definition QAs, we generated two close-up CAD images (like that in Figure \ref{fig:cad_rep}B), one which contained the component and another which did not. These 62 close-up CAD images were appended to the corresponding multi-view CAD image (like that in Figure \ref{fig:cad_rep}A), which provided more 3D context for the close-up image. This resulted in 62 VQA pairs, each of which had the following prompt, where $\{component\_name\}$ was replaced with one of the 31 components:

\begin{customquote}
Also attached is an image showing seven CAD views (each boxed in black) of our vehicle design. The top, big view shows a close-up view of the design. The six smaller views on the bottom of the image show different complete views of the CAD of the vehicle and are provided for context. Note that the close-up view orientation matches one of the six complete view orientations. The close-up view may also have some components hidden (with respect to the corresponding complete view) for visualization of specific components. Looking at the close-up view, is/are the $\{component\_name\}$ visible in the close-up view? Answer simply with yes or no.
\end{customquote}

For the Definition and Presence questions, we also tracked how the component-in-question was mentioned in the rule document: if it was mentioned explicitly in a ``definition'' section of the FSAE rule document (``definition component''), if it was not in a definition section but mentioned multiple times throughout the document (``multi-mention component''), or if it was not mentioned in the document at all (``no-mention component''). The intent behind this tracking was to understand whether the frequency and way in which a component's name is mentioned in the rule document is correlated with the model's ability to visually identify the component in the Rule Comprehension questions.

\subsubsection{Rule Compliance}
\label{sec:rule_compliance}
Engineers frequently consult requirement documents to ensure that their designs comply with specific specifications. The Rule Compliance segment of the benchmark characterizes a model's ability to check that a design conforms with a specific rule. This part of the benchmark is further divided into two subsets: Dimension QAs and Functional Performance QAs, depending on the type of rule in question.

\textbf{Dimension} questions test a model's ability to check that a design complies with a rule that stipulates dimensional constraints. From a list of 20 dimension rules, we generated three engineering drawing images for each rule:
\begin{enumerate}
    \item A direct-dimensioned and rule-compliant image (as in Figure \ref{fig:cad_rep}C).
    \item A direct-dimensioned and rule-violating image. These were generated by editing the dimensions on the first image to explicitly violate the rule-in-question, or by modifying the CAD model so that the updated dimensions violated the rule.
    \item A scale-bar-dimensioned and rule-compliant image (as in Figure \ref{fig:cad_rep}D).
\end{enumerate}
No scale-bar-dimensioned and rule-violating QAs were created. Since the CAD provided by MIT Motorsports is inherently rule-compliant, it is difficult to create negative examples when editing of direct-dimensions is not possible. Each of these engineering drawing images was appended to a corresponding multi-view CAD image (like that in Figure \ref{fig:cad_rep}A) to provide context about the full model. This resulted in 60 VQAs with the following prompt, appended to the preamble:

\begin{customquote}
Also attached is an image that shows an engineering drawing of the vehicle on the top accompanied by six CAD views of the vehicle on the bottom. The six CAD views each feature a different orientation of our design, so that 3D information about our design can be inferred. The CAD views are provided to contextualize the engineering drawing, which has the same orientation as one of the six CAD views. All units displayed in the engineering drawing have units of mm. Based on the engineering drawing, does our design comply with rule $\{rule\_number\}$ specified in the FSAE rule document? 

\{[\textit{If direct-dimensioned:}] Only use dimensions explicitly shown in the engineering drawing to answer the question. If a dimension is not explicitly shown, you can assume that it complies with the rules.\}

\{[\textit{If scale-bar-dimensioned:}] To answer the question, use the scale bar shown at the top of the engineering drawing to compute necessary dimensions in the drawing.\}

First provide an explanation for your answer (begin it with 'Explanation:'). Then provide just a yes/no answer (begin it with 'Answer:') that summarizes your response.
\end{customquote}

While these questions can be answered with a yes/no response, we also wanted to assess the model's ability to explain \textit{why} the design was or was not compliant, encouraging chain-of-thought reasoning \cite{lu2022learn}. For each direct-dimensioned question, we (or members of the MIT Motorsports team) wrote an explanation justifying the ground-truth yes/no answer. These explanations were not extensively reviewed, other than to ensure that they supported the corresponding ground truth yes/no answer. These human-written explanations can then be compared to generated model explanations.

From these 60 VQAs, we generated another set of 60 VQAs with additional context that would help with answering the question. For the set of original 60 VQAs, we swapped out the multi-view CAD portion of the image with a different multi-view CAD image with components highlighted in pink that were relevant to the rule. We also added a line to the prompt explaining what the highlighted components were (e.g. ``In the CAD views, the lower side impact structure is highlighted in pink''). In total, 120 Dimension VQAs were generated: 60 without additional context and 60 with additional context.

\textbf{Functional Performance} questions also test a model's ability to check that a design complies with a rule given a relevant image. For this category however, either the rule, the image, or both is related to some functional performance of the design. Most questions involve a rule that imposes a constraint on some functional criteria of the design, and the image conveys the information required to check the rule. For example, there could be a restriction on the material choice for a part (hence the corresponding material strength) in the rule and the visualization of FEA results could indicate the maximum stress found in the part. When applicable, a pair of positive and negative examples is generated, where a variation is introduced to either the question or the image such that the first example would violate the rule while the second example would not. There is significant variation in the images and rules contained within this subset, and thus there is no standardized prompt. We encourage exploration of our code for more details. As these questions were more difficult to formulate due to limited availability of functional performance data, this subset has 16 VQAs. Like the Dimension questions, we generated a human-written explanation for each VQA.

\subsection{Evaluation Metrics}
\begin{figure*}[!ht]
\begin{center}
\setlength{\unitlength}{0.012500in}%
\includegraphics[width=\linewidth]{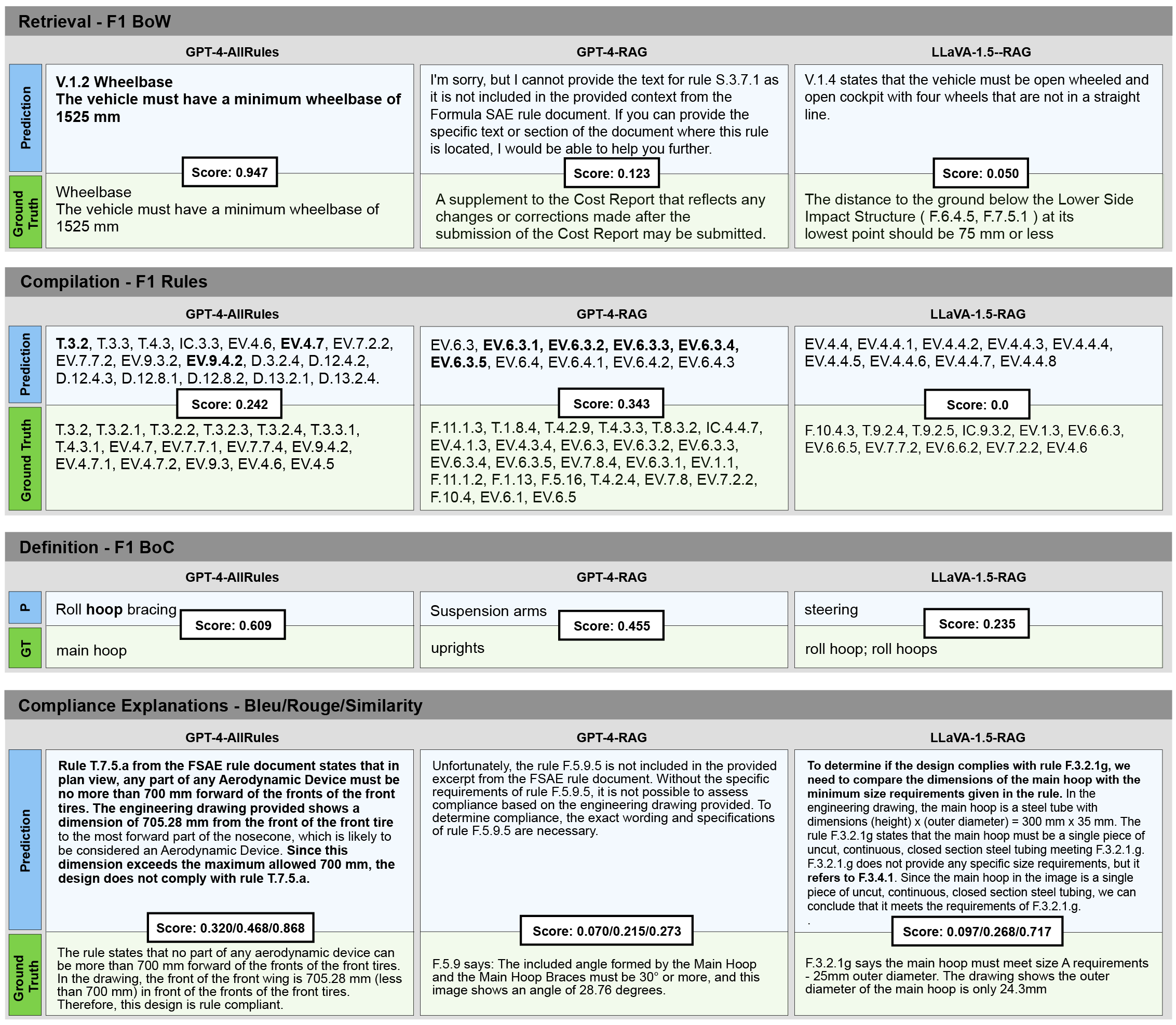}
\end{center}
\caption{Sample responses from GPT-4-AllRules, GPT-4-RAG, and LLaVA-1.5-RAG across different subsets of the benchmark. We show the subsets that have evaluation metrics that can be harder to interpret, to provide references for various scores. The bolded portions of the predicted responses show what we interpreted to be correct.}
\label{fig:model_responses} 
\end{figure*}

Models tested on \acr{} can be evaluated completely automatically. For each of the six subsets of the benchmark, an appropriate automated evaluation metric was selected and implemented in our code so that a model's (predicted) answer can readily be compared to the ground-truth answer. For several of the evaluation metrics, it can be difficult to intuit what score a prediction would receive relative to a ground-truth answer. As such, we've provided some examples in Figure \ref{fig:model_responses} of model predictions (from our Model Evaluation section), their corresponding ground-truth answers, and the resulting scores. Each of the evaluation metrics is discussed in-depth below.

\subsubsection{F1-score}
Many subsets of our dataset are scored on a flavor of the F1-Score. The F1-Score is a popular metric for evaluating models' performance on binary classification tasks, as the metric weighs both precision and recall. It is defined as:  \( \text{F1-Score} = 2 \times \frac{\text{Precision} \times \text{Recall}}{\text{Precision} + \text{Recall}} \).

\textbf{F1 Bag of Words:} F1 applied to a Bag of Words (BoW) was used by~\cite{rajpurkar2016squad, yang2015wikiqa, dasigi2021dataset} as an automatic metric for their benchmarks, which asked a model to pull verbatim phrases from a body of text to answer a question. As defined by~\cite{rajpurkar2016squad}, the metric first involves a cleaning step. The predicted answer (model response) and the ground-truth answer are converted to lower-case characters, extra white-space is removed, and punctuation and articles are taken out. They are then each ``tokenized'' into lists of words -- predicted list (P) and ground-truth list (GT) -- where F1-Score can be computed using Precision and Recall, where \(Precision = \frac{P \cap GT}{\text{len}(P)}\) and \(Recall = \frac{P \cap GT}{\text{len}(GT)}\).

Since our Retrieval QAs also ask the model to pull text from the rule document verbatim, we use this F1 Bag of Words metric to evaluate the Retrieval subset of the benchmark. We compute F1 BoW for each QA, and we report a macro-average across all questions.

\textbf{F1 Rules:} Similar to the Retrieval QAs, our Compilation questions ask a model to identify text (specifically rule numbers) contained within the rule document. We therefore use a very similar metric to the F1 Bag of Words used for the Retrieval QAs, except the lists P and GT are replaced by lists of rule numbers. We compute F1 Rules for each QA, and we report a macro-average across all questions.

\textbf{F1 Bag of Characters (BoC):} Our Definition QAs ask a model to identify a component highlighted in a multi-view CAD image, using the rule document for reference. These QAs seemed like they should be scored similarly to the Retrieval QAs; however, since the model was now being asked for component names (several words) rather than complete rules (sentences), we did not want to penalize the model for small spelling errors or ending differences. For example, if the ground truth is ``front hoop,'' a predicted response of ``front hoope'' should be considered more correct than ``front motor.'' F1 Bag of Words would score ``front hooped'' and ``front motor'' as equally correct. F1 Bag of Characters reflects the relative correctness of ``front hooped'' over ``front motor.'' It is computed in the same way as F1 Bag of Words, except tokenization occurs on the character rather than word level. We compute the F1 BoC for each Definition QA across all synonyms, and we report the macro average of the highest score for each QA.

\subsubsection{Accuracy} Several subsets of our benchmark -- Presence, Dimension, and Functional Performance questions -- ask the model to provide a yes/no answer. We score these using accuracy (ACC).

\subsubsection{Explanation Metrics}
For the Rule Compliance questions, models return explanations as well as yes/no answers. Automatically scoring model-generated explanation text for semantic similarity to a human-written reference text is a complex task, with no perfect solution. A recently developed benchmark testing large language models on multiple-choice science questions, ScienceQA~\cite{lu2022learn}, used three different automated metrics -- BLEU~\cite{papineni2002bleu}, ROUGE~\cite{lin2004rouge}, and Similarity~\cite{reimers2019sentence} -- to assess similarity between LLM-generated and human-written explanations. We follow Lu et al.'s example and use these three metrics to score the MLLM-generated explanations relative to the human-collected ones in our benchmark. Each of these metrics is described further below.

\textbf{BLEU:}
The BLEU (Bilingual Evaluation Understudy) score was developed by Papineni et al.~\cite{papineni2002bleu} for the automatic scoring of machine translations relative to human, reference translations. To compute BLEU, predicted and reference sentences are first broken up into n-grams, which are segments of n (a user-specified number) words. n-gram matches between the predicted and reference sentences are then found; once a predicted n-gram is matched to a reference n-gram, the n-gram is removed from the pool of reference n-grams in a process called ``clipping.'' The number of matching n-grams is then divided by the number of n-grams in the predicted sentence, producing a modified precision score, $p_n$. The authors suggest to compute BLEU-4: $p_n$ for n = 1 through n = 4, taking the geometric mean of the four $p_n$ (since $p_n$ decays exponentially with increasing n)~\cite{papineni2002bleu}. However, since we only use one reference (one explanation), BLEU-4 scores were always near zero. As such, we report BLEU-2, which has non-zero scores but still preserves some of the information about adjacency of words.
BLEU-2 (with max n-gram 2) can be computed as:
\( \log(\text{BLEU}) = \min\left(1 - \frac{r}{c}, 0\right) + \sum_{n=1}^{2} \frac{1}{2} \log(p_n) \)

The min() term is a ``brevity penalty'' and serves to penalize predictions (of length c) that are shorter than the ground truth (length r). For the explanation portions of the Rule Compliance questions, we report BLEU-2 to quantify the similarity between the model's generated explanation and the human-generated explanation.

\textbf{ROUGE:}
The ROUGE (Recall-Oriented Understudy for Gisting Evaluation) metric was developed by Lin et al.~\cite{lin2004rouge} to measure the quality of a computer-generated text summary relative to a human, reference summary. ROUGE-L, which uses longest common sub-sequence (LCS), is useful for characterizing similarity in sentence-level word order. The LCS refers to the longest sequence of words that appear in the same order but not necessarily consecutively in the two sentences. ROUGE-L is computed as the F1 score of the LCS:
\(R_{LCS} = \frac{LCS(R, P)}{m}\), \(P_{LCS} = \frac{LCS(R, P)}{n}\), and \(ROUGE_L = \frac{2P_{LCS} \times R_{LCS}}{P_{LCS} + R_{LCS}}\), 

where $R$ is a reference sentence, $P$ is a predicted sentence, $m$ is the length of a reference sentence, $n$ is the length of a predicted sentence, $R_{LCS}$ is the recall, and $P_{LCS}$ is the precision. For the explanation portions of the Rule Compliance questions, in addition to reporting BLEU-2, we also report ROUGE-L.

\textbf{Similarity:}
The Similarity metric refers to Sentence-BERT-based sentence embedding cosine similarity~\cite{reimers2019sentence, zhang2019bertscore}. Both the model-generated explanation and the human-written explanation are embedded using the \textit{sentence-transformers/all-MiniLM-L6-v2} sentence transformer, resulting in two vectors that encode the meaning of the explanations. Cosine similarity between the two vectors is then computed to identify how much overlap there is in meaning between the two explanations.

\section{MODEL EVALUATION}

We evaluate simple baselines and state-of-the-art MLLMs, at the time of writing, on our \acr{} benchmark to understand current AI models' capabilities in understanding engineering requirement documentation. The goal of these evaluations is not to be exhaustive: there are many other models that should be tested, time and cost permitting. Given the rapid pace of MLLM development, we ask for the research community's assistance in benchmarking new state-of-the-art models as they are developed. The goal of the evaluations we conducted is rather to illustrate the gaps in current AI's capabilities and encourage other researchers to build and train better AI models and frameworks that target these gaps. Moreover, evaluating different MLLM models provides some insight into the relative difficulty of the questions, and the failure modes could serve as inspiration for future approaches to benchmarking design-requirement-type questions. 

\subsection{Baselines and Models}
\label{sec:baselines_models}

\subsubsection{Naive Baselines}
Similar to \cite{shaham2023zeroscrolls}, we create basic baselines that rely on random selection so that it is easier to contextualize models' performances across the different subsets and metrics of the benchmark. For the Retrieval questions, we randomly choose a rule from the 1192 rules in our rule list. For the Compilation questions, we randomly pick 10 rules from our list of 1192. For the Definition questions, we randomly choose two consecutive words in the rule document. For the questions scored on accuracy (Presence, Dimension, and Functional Performance questions), we randomly select yes or no with 50\% probability. Note that these baselines are evaluated to provide a sense of the lower threshold score (predicting randomly with no learning) for any machine learning model.

\subsubsection{MLLM Models}
\label{sec:mllm_models}
We consider five recently developed MLLMs in our evaluation: four closed-source models -- OpenAI's \textit{gpt-4o} (GPT-4o) \cite{gpt4o}, OpenAI's \textit{gpt-4-1106-vision-preview} (GPT-4) \cite{gpt4v}, Google AI's \textit{models/gemini-1.0-pro-vision} (Gemini-1.0) \footnote{https://ai.google.dev/gemini-api/docs/models/gemini?authuser=1}, and Anthropic's \textit{claude-3-opus-20240229} (Claude-Opus) \footnote{https://www.anthropic.com/news/claude-3-family} -- and one open-source model: \textit{llava-1.5-13b} (LLaVA-1.5) \cite{liu2023llava}. Note that all future references to GPT-4o, GPT-4, Gemini-1.0, Claude-Opus, or LLaVA-1.5 refer to these specific model versions unless otherwise stated. GPT-4, Gemini-1.0, and Claude-Opus were originally selected for their strong performances as closed-source models on existing multimodal benchmarks \cite{shaham2023zeroscrolls, picard2023concept}. For example, Gemini-1.0 Ultra, Claude-Opus, and GPT-4V(ision) were the top three models, as of April 2024, on the MMMU benchmark, which consists of 11.5k multimodal questions from college curriculum content \cite{yue2023mmmu}. GPT-4o was released towards the end of the time of this work. It holds the 1st place spot on the MMMU leaderboard \cite{yue2023mmmu}, as of August 2024. It was added to our closed-source model evaluations both because of this notable performance and because of potential for comparison with GPT-4. In this way, DesignQA can be used to continue to measure progress of models as they develop over time. LLaVA-1.5 is selected because of its promise (and the promise of its derivatives) as an open-source MLLM~\cite{liu2023improved}. While the Retrieval and Compilation questions in the benchmark could be answered with text-only models, we pose the questions in this segment with a null image to the MLLMs.

The extracted text from the FSAE rule document PDF is roughly 70,091 tokens in length. As GPT-4o and GPT-4 have 128,000 token context windows and Claude-Opus has a 200,000 token context window, these models can ingest the whole rule text in the prompt. However, LLaVA-1.5 and Gemini-1.0, which only have 4,096 token and 12,288 token context windows respectively, cannot take the whole rule text as context input. Thus, these two models require a Retrieval Augmented Generation (RAG) system to retrieve the appropriate context relevant to the question from the FSAE rule document~\cite{lewis2020retrieval}. As our research goal was not to evaluate different RAG techniques, we implement a very simple RAG using the LlamaIndex framework, described in more detail in Section~\ref{sec:rag}.  In the cases of GPT-4o, GPT-4, and Claude-Opus, ingesting the whole rule document via the context window proved to be costly, ten times or more the cost of providing the rule document to the model via the simple RAG. Of these models, Claude-Opus had the most expensive cost per input token at the time of writing, so we only tested Claude-Opus with a RAG system. For GPT-4o and GPT-4, we tested the models by providing them the full rule document via their context windows and also by providing them the rule document via RAG, so as to enable better comparison with the other models that received the rules via RAG.

In summary, we tested the following model combinations: GPT-4o-AllRules (given the full rule document via its context window), GPT-4-AllRules (given the full rule document via its context window), GPT-4o-RAG, GPT-4-RAG, LLaVA-1.5-RAG, Gemini-1.0-RAG, and Claude-Opus-RAG. For all models, we did not specify a system prompt template, so the default system prompt was used. We also used the default parameters (top\_p and temperature) for each model, as set by LlamaIndex, the framework from which we called the MLLMs. More details can be found in the SI.

\subsubsection{RAG}
~\label{sec:rag}
RAG is a natural language processing technique that enhances text generation by incorporating external knowledge, dynamically retrieving relevant information from a database or corpus to inform and improve the content being produced~\cite{lewis2020retrieval}. It is especially helpful for models with small context windows, as it selectively pulls relevant information from an extensive document to assist in generating more informed and contextually accurate text outputs. As mentioned before, our goal in this work is not to study different RAG techniques. As such, we implemented a very simple RAG system using LlamaIndex and using OpenAI's \textit{text-embedding-3-large} to embed the information in the FSAE rule document~\cite{Liu_LlamaIndex_2022}. We index the text from the rule document into 250-token chunks, with a 50-token overlap. From the embeddings, the cosine similarity between the question and each of the embedded chunks is then computed. The top-15 (top-12 for Compliance QAs) most relevant pages are then fed into the prompt as context, before posing the question in the benchmark. If including the top-15 most relevant pages exceeded the model's context window size, the top-k value was lowered. While there are a few images in the FSAE rule document, they were not considered for our evaluation or included in our RAG, as they require further processing, and LLaVA-1.5 was not trained on multiple image inputs. The few tables in the FSAE rule document will not receive any special treatment and will be fed into the models just as simple text from the PDF text-extracting script, together with the rest of the text on the page.

For just the Definition question subset of the benchmark, the *-RAG models were not given any RAG-generated context. This is because the questions in the Definition subset are the same, i.e., ``tell me the name of the highlighted component,'' even though the images vary across QAs. As such, RAG always returned the same portion of the rule document for each question. Since this identical context is unhelpful, we did not provide it to the models for the Definition questions. As such, all *-RAG models tested did not receive document context just for the Definition subset.

\subsection{Results and Analysis}

Table~\ref{tab:results} shows all the baseline and model results. First, we discuss some overall findings and then we specifically delve into the results for each subset of the benchmark.

\paragraph{Overall Results}
Of the models tested, GPT-4o-AllRules performs the best across almost all metrics in the benchmark. GPT-4o-AllRules and GPT-4-AllRules, in all but one subset case, perform better than their corresponding -RAG models, demonstrating that the simple LlamaIndex RAG implementation was generally ineffective at providing relevant rule information to the models. In contrast, inputting the full rule text into the context window allowed GPT-4o-AllRules and GPT-4-AllRules to access necessary information. Of the models tested, GPT-4o-AllRules is not the best performer on the Rule Compliance explanations: Gemini-1.0-RAG receives higher scores (or ties) on the BLEU and ROUGE metrics, likely because of the model's succinct explanations, and Claude-Opus-RAG ties GPT-4o-AllRules on the Functional Performance explanation Similarity score. Of the *-RAG models tested, GPT-4o-RAG tends to be the best performer, with the exception of the Functional Performance questions, where Claude-Opus-RAG has the highest Functional Performance accuracy score of the *-RAG models tested. This finding perhaps indicates that Claude-Opus has better engineering-design technical reasoning skills than the other models evaluated.

Despite GPT-4o-AllRules's notable performance of the models tested, it does not attain perfect scores (1.0 for each subset of the benchmark). Even though it receives the full rule document, GPT-4o-AllRules cannot reliably extract specific rules from the text, a trivial yet critical task for engineers. MLLMs' scores on the Compilation, Definition, and Presence subsets of the benchmark also remain relatively low when compared with the Naive baseline, possibly because these subsets' questions don't point the models to specific sections of the rule document, resulting in harder search problems. There remains significant work that must be done to  develop MLLMs equipped with engineering and reasoning knowledge that can solve these problems characteristic of real-world engineering design challenges. 
We must focus efforts on developing MLLM methods that excel in engineering contexts, enhancing precision and reliability in real-world applications.

\begin{table*}[th]
\centering
\caption{This table presents a detailed comparison of various MLLM models' scores on our benchmark. Of the models tested, GPT-4o-AllRules is the best performer across nearly all metrics. The Compilation, Definition, and Presence questions have the smallest difference between the best performing model's score and the Naive baseline's score. For all metrics, a perfect score would be 1.0 and} larger values are better, as indicated by the up arrows.
\label{tab:results}
\resizebox{\textwidth}{!}{%
\begin{tabular}{@{}lccccccccc@{}}
\toprule
\multirow{1}{*}{\makecell[l]{\textbf{Subset}\\\textbf{(Metric)}}} & \textbf{Baseline} & \multicolumn{7}{c}{\textbf{Model}} \\ 
\cmidrule(lr){2-2} \cmidrule(lr){3-9}
& \textbf{Naive} &\textbf{GPT-4o-AllRules} &\textbf{GPT-4-AllRules} &\textbf{GPT-4o-RAG} & \textbf{GPT-4-RAG} & \textbf{LLaVA-1.5-RAG} & \textbf{Gemini-1.0-RAG}  & \textbf{Claude-Opus-RAG} \\ 
\midrule
RULE EXTRACTION & & & & & & & & \\
\hline
\addlinespace
\makecell[l]{\textbf{Retrieval}\\(F1 BoW $\uparrow$)} & 0.082 & \textbf{0.885} & 0.750 & 0.186 & 0.181 & 0.112 & 0* & 0.173 \\
\addlinespace
\makecell[l]{\textbf{Compilation}\\(F1 rules $\uparrow$)} & 0.137 & \textbf{0.424} & 0.298 & 0.376 & 0.362 & 0.281 & 0.283 & 0.288 \\ 
\addlinespace
\midrule
RULE COMPREHENSION & & & & & & & & \\
\hline
\addlinespace
\makecell[l]{\textbf{Definition}\\(F1 BoC $\uparrow$)}& 0.358 & \textbf{0.540} & 0.470 & 0.525 & 0.420 & 0.393 & 0.488 & 0.423\\
\addlinespace
\makecell[l]{\textbf{Presence}\\(ACC $\uparrow$)} & 0.5 & \textbf{0.726} & 0.629 & 0.710 & 0.532 & 0.484 & 0.548 & 0.5 \\ 
\addlinespace
\midrule
RULE COMPLIANCE & & & & & & & & \\
\hline
\addlinespace
\makecell[l]{\textbf{Dimension}\\(ACC $\uparrow$)\\(BLEU/ROUGE/Similarity $\uparrow$)} & \makecell{\\0.5\\-} & \makecell{\\\textbf{0.825}\\0.175/\textbf{0.344}/\textbf{0.777}} & \makecell{\\0.533\\0.118/0.296/0.728} & \makecell{\\0.675\\0.111/0.255/0.642} & \makecell{\\0.300\\0.091/0.235/0.592} & \makecell{\\0.408\\0.097/0.241/0.578} & \makecell{\\0.525\\\textbf{0.176}/\textbf{0.344}/0.644} & \makecell{\\0.508\\0.137/0.295/0.698}\\
\addlinespace
\makecell[l]{\textbf{Functional Performance}\\ (ACC $\uparrow$)\\(BLEU/ROUGE/Similarity $\uparrow$)} & \makecell{\\0.5\\-} & \makecell{\\\textbf{0.938}\\0.230/0.408/\textbf{0.745}} & \makecell{\\0.563\\0.167/0.342/0.697} & \makecell{\\0.750\\0.181/0.367/0.736} & \makecell{\\0.563\\0.121/0.306/0.698} & \makecell{\\0.536\\0.163/0.321/0.650} &  \makecell{\\0.438\\\textbf{0.266}/\textbf{0.444}/0.725} & \makecell{\\0.875\\0.172/0.354/\textbf{0.745}}\\ 
\bottomrule
\end{tabular}%
}
\footnotesize{*Gemini-1.0-RAG received a score of 0 on the Retrieval question because of a ``RECITATION'' error that caused it to stop response generation.}
\end{table*}

\paragraph{Rule Extraction: Retrieval} 
Of the models evaluated, GPT-4o-AllRules and GPT-4-AllRules significantly outperform the other models at retrieving the requested rule verbatim (0.885 and 0.750 average F1 BoW score, respectively). When an *-AllRules model's answer diverged from the ground truth answer, we noticed that it was often because the model was reporting a nearby rule rather than the rule requested (e.g. V.3.2.5 instead of V.3.2.4) or because the model was reporting a similarly numbered rule (e.g., F.3.2.1) instead of the requested rule (e.g., V.3.2.1). Sometimes the *-AllRules models' answers were different from the ground truth because they included all child rules in addition to the requested parent rule (e.g., when asked for V.1, V.1 was reported along with V.1.1 and V.1.2). While not technically wrong, this result was not handled differently by our evaluation metric and may be a result we would want to handle specifically in the future.

Providing the rules to GPT-4o or GPT-4 via RAG instead of via context resulted in a much lower average F1 BoW score (0.186 and 0.181, respectively).  For a number of questions (like the GPT-4-RAG Retrieval example shown in Figure \ref{fig:model_responses}), GPT-4o-RAG and GPT-4-RAG replied that they could not produce the text for the rule because it was not included in the rule text given to it, indicating that the simple LlamaIndex RAG failed to provide the relevant portion of the rule document to the models. When GPT-4o-RAG and GPT-4-RAG did provide an answer to the question, they had similar responses to that of their respective *-AllRules models. Claude-Opus-RAG had a similar average F1 BoW score (0.173) to that of GPT-4o-RAG and GPT-4-RAG, and Claude-Opus-RAG also noted when the rule was not present in the provided RAG context.

LLaVA-1.5-RAG received the same portion of the rule document as the other *-RAG models, but its average F1 BoW score was even lower (0.112). Unlike GPT-4o-RAG, GPT-4-RAG, and Claude-Opus-RAG, LLaVA-1.5-RAG would hallucinate rules rather than indicate that the requested rule was not contained within the portion of the document it received via RAG. Furthermore, instead of returning rules verbatim, LLaVA-1.5-RAG would frequently offer an interpretation of the requested rule (as in the LLaVA-1.5-RAG Retrieval example in Figure \ref{fig:model_responses}).

Gemini-1.0-RAG received a score of zero on the Retrieval subset of the benchmark because the model encountered a ``RECITATION'' error for each question. The model has been programmed to stop content generation if the generated content repeats parts of the input data. While in some cases this behavior is useful, it is not useful for the purpose of this subset of the benchmark where we request verbatim repetition. We can imagine many other similar queries -- where direct quotation is desired by a user -- for which this repetition blocking would be problematic. Since this is a closed-source model, there is no way for us to alter this behavior.

\paragraph{Rule Extraction: Compilation} 
GPT-4o-AllRules performs the best (0.424 F1 rules) on the Compilation questions of the models tested. However, the Compilation questions are one of the benchmark categories (along with the Definition and Presence questions) which has a relatively small difference between the best performing model score (GPT-4o-AllRules) and the naive random baseline score. What inhibits GPT-4o-AllRules -- the best performer -- from having a higher score? We noticed that the model repeatedly fails to include rules in its answer that contain the search term verbatim. For example, in the first Compilation question, which asks the model for rules that pertain to ``aerodynamic or aerodynamics,'' the model neglects to include T.7.2.1, T.7.2.2, and T.7.5, which all include ``aerodynamic'' verbatim in their rule text. Improving MLLMs' basic text extraction skills is essential to their performing better on design-requirement-type tasks. We also observed that GPT-4o-AllRules frequently neglects to report rules that are mentioned by rules that include the search term. For example, in the aerodynamic question, GPT-4o-AllRules doesn't mention GR.6.4.1, V.1.1, and V.1.4.1, all of which are referenced by other rules that contain the word ``aerodynamic.'' While perhaps our prompt should explicitly state that rules referred to by rules that contain the search term are relevant, MLLMs must have the capability to follow chains of rules across disconnected portions of a document.

In terms of the *-RAG models, GPT-4o-RAG and GPT-4-RAG performed similarly on the Compilation questions, while the other *-RAG models didn't perform as well. Claude-Opus-RAG's lower score (compared with GPT-4o-RAG's) could be attributed to its failure to follow the provided instructions. The prompt specifically asked for a list rules separated by commas, using no other words in the response. Claude-Opus-RAG includes other words in each of its answers, starting each answer with ``The rules relevant...'' and occasionally separating rules by spaces instead of commas. Since automated evaluations were used, this lack of instruction following resulted in lower scores. Gemini-1.0-RAG and LLaVA-1.5-RAG also had lower scores (0.283 and 0.281, respectively) than GPT-4o-RAG. The models' answers to 6/30 (Gemini-1.0-RAG) and 5/30 (LLaVA-1.5-RAG) of the questions had a score of zero (no overlap with the ground truth rule list). In contrast, GPT-4o-RAG had a score of zero for just one out of the 30 questions.

\begin{table*}[th]
\centering
\caption{Results from the GuaranteedRAG experiment. We compare LLaVA-1.5's performance on the benchmark when given FSAE rule document sections via smiple LlamaIndex RAG and when given GuranteedRAG.}
\label{tab:idealrag_results}
\begin{tabular}{@{}lcc@{}}
\toprule
\multirow{1}{*}{\makecell[l]{\textbf{Subset}\\\textbf{(Metric)}}} & \multicolumn{2}{c}{\textbf{Model}} \\ 
\cmidrule(lr){2-3}
& \textbf{LLaVA-1.5-RAG} & \textbf{LLaVA-1.5-GuaranteedRAG}  \\ 
\midrule
RULE EXTRACTION & &\\
\hline
\addlinespace
\textbf{Retrieval} (F1 BoW $\uparrow$) & 0.112 & \textbf{0.699 $\pm$ 0.00590} \\
\textbf{Compilation} (F1 rules $\uparrow$) & 0.281 & \textbf{0.357 $\pm$ 0.0205}\\ 
\midrule
RULE COMPREHENSION & & \\
\hline
\addlinespace
\textbf{Definition} (F1 BoC $\uparrow$) & 0.393 & \textbf{0.471 $\pm$ 0.0255}\\
\textbf{Presence} (ACC $\uparrow$) & 0.484 & \textbf{0.543 $\pm$ 0.0353}\\ 
\midrule
RULE COMPLIANCE & &\\
\hline
\addlinespace
\makecell[l]{\textbf{Dimension} (ACC $\uparrow$)\\(BLEU/ROUGE/Similarity $\uparrow$)} & \makecell{0.408\\0.097/0.241/0.578} & \makecell{\textbf{0.482 $\pm$ 0.0473}\\\textbf{0.121 $\pm$ 0.008/0.266 $\pm$ 0.0105/0.652 $\pm$ 0.0158}}\\ 
\addlinespace
\makecell[l]{\textbf{Functional Performance} (ACC $\uparrow$)\\(BLEU/ROUGE/Similarity $\uparrow$)} & \makecell{\textbf{0.536}\\\textbf{0.163/0.321/0.650}} &  \makecell{0.375 $\pm$ 0.159\\0.152 $\pm$ 0.0165/0.316 $\pm$ 0.0269/0.650 $\pm$ 0.0154}\\ 
\bottomrule
\end{tabular}%
\end{table*}

\paragraph{Rule Comprehension: Definition}
Of the models tested, GPT-4o-AllRules performs the best on the Definition subset of the benchmark. The Definition questions had the smallest difference between the best-performing model score and the naive random baseline score of any subset of the benchmark, indicating that they are challenging for MLLMs to answer. Contributing to GPT-4o-AllRules's relatively low score on the Definition questions were visual component recognition errors. For example, when shown independent images where the steering wheel, steering column, and steering rack were each highlighted in pink, GPT-4o-AllRules never mentioned something related to ``steering'' as the highlighted component in-question. Instead, it answered ``impact attenuator'' twice and ``front hoop'' once. These components are very different parts of the vehicle from the steering system. As such, visual component recognition capabilities, especially when it comes to technical components, needs to be improved for MLLMs. Without good component understanding, MLLMs will struggle to answer Rule Compliance questions that pertain to those components.

\paragraph{Rule Comprehension: Presence}
GPT-4o-AllRules performed the best (0.726 accuracy) on the Presence questions of the models tested. The *-RAG models all had lower average accuracies likely because these models had access to limited (15 or fewer) rule document pages that reference the component in question. Notably, Claude-Opus-RAG and LLaVA-1.5-RAG perform equivalent to and worse than the naive baseline, respectively. For 28/30 of the questions, Claude-Opus-RAG answered ``no'' to the posed question, perhaps reflecting a cautious stance. Since only half of the questions had a ground truth answer of no, consistently answering no amounts to the same as randomly guessing (the naive baseline). LLaVA-1.5-RAG had no noticeable trends in yes/no prediction, but performed worse than random choice.

\paragraph{Rule Compliance: Dimension}
Of the models tested, GPT-4o-AllRules performs the best in terms of accuracy on the Dimension questions. It performs well, with an accuracy score of 0.825. We looked closer at where GPT-4o-AllRules struggled in answering the basic (no additional context, see Section \ref{sec:rule_compliance}) Dimension questions, to gain insight into its failure modes. For five questions that it answered incorrectly, GPT-4o-AllRules struggled to compute accurate dimensions using a scale bar. Models' understanding of scale bar versus directly dimensioned engineering drawings is explored further in Section \ref{sec:discussion}. Two of the questions that GPT-4o-AllRules answered incorrectly featured engineering drawings that displayed two dimensions; the correct answer to the question required that the model take a sum or difference between the two dimensions. The model struggled on these questions because it scraped a single dimension as the answer, rather than scraping both dimensions and combining them to get the answer. In another two questions GPT-4o-AllRules answered incorrectly, it mistakenly identified which vehicle components the dimensions were spanning in the engineering drawings.  We also observed an instance of the model extracting an incorrect number from the engineering drawing image (9 instead of 0.9) and an instance of the model referencing the wrong rule from the rule document. These findings indicate that MLLMs' still have room for improvement when it comes to skills -- like visual component recognition, OCR, and dimensional computation -- that are needed to answer questions pertaining to engineering drawings.

Of the models tested, GPT-4o-AllRules has the highest Similarity score for its Dimension explanations, matching its best-performer accuracy score. However, it is not the best-performer on the BLEU and ROUGE metrics, where Gemini-1.0-RAG has the higher (or tied, in the case of ROUGE) scores. We observe that many of the MLLMs' explanations are significantly longer than the reference explanations we collected (see Figure \ref{fig:model_responses}). We suspect that Gemini-1.0-RAG's high scores on the BLEU and ROUGE metrics are because its explanations tended to be shorter, more comparable in length to the human-generated, reference explanations. It is therefore challenging to capture the correctness of explanations through the BLEU and ROUGE scores. The interpretability of these scores would be improved by obtaining more human reference explanations, so as to better capture the distribution of possible explanations that could be considered correct.

\paragraph{Rule Compliance: Functional Performance}
GPT-4o-AllRules once again performed the best on this subset of the benchmark, of the models tested. Its accuracy score was quite high, answering only one of the 16 questions incorrectly. The question the model answered incorrectly pertains to an anthropomorphic data chart showing requirements for human size (height, weight, etc.) for the vehicle and another chart that shows the human sizes the built vehicle can actually accommodate. The model only compares one row of each chart (heights) before deciding that the vehicle is in compliance, when the weight row of the vehicle's actual capabilities violates the corresponding weight row of the requirement chart.

When comparing the *-RAG models, we find that, in a departure from other benchmark subsets where GPT-4o-RAG is the best performer, Claude-Opus-RAG performs the best on Functional Performance accuracy, with a score of 0.875. The model also ties GPT-4o-AllRules's high score on explanation Similarity, further underscoring its good understanding of these questions. This subset of the benchmark required significant technical expertise, perhaps indicating that Claude-Opus has one of the stronger engineering backgrounds of the MLLMs tested. It would be interesting to test and compare Claude-Opus-AllRules to GPT-4o-AllRules, but given the cost constraints of running Claude-Opus-AllRules mentioned previously, this will be left as future work.

\begin{table*}[th]
\centering
\caption{Analysis of MLLMs' performances on the Rule Comprehension questions, broken-down by vehicle-component-mention type. Def refers to definition-components, those defined explicitly within the rule document. MultM refers to multi-mention components, those that appear multiple times within the rule document but are not explicitly defined. NoM refers to no-mention components, those that don't appear verbatim in the rule document. Trends appear in that all models score the highest on definition-components questions within the Definition subset, while all models score the highest on the multi-mention component questions within the Presence subset.}
\label{tab:definition_mentions}
\begin{tabular}{@{}llccccccc@{}}
\toprule
 \multirow{3}{*}{\textbf{Subset}}  &  \multirow{4}{*}{\makecell[l]{\textbf{Mention}\\\textbf{Type}}} & \multicolumn{7}{c}{\textbf{Model}} \\ 
\cmidrule(lr){3-9}
& & \makecell{\textbf{GPT-4o-}\\\textbf{AllRules}} & \makecell{\textbf{GPT-4-}\\\textbf{AllRules}} & 
\makecell{\textbf{GPT-4o-}\\\textbf{RAG}} &
\makecell{\textbf{GPT-4-}\\\textbf{RAG}} & \makecell{\textbf{LLaVA-}\\\textbf{1.5-}\\\textbf{RAG}} & \makecell{\textbf{Gemini-}\\\textbf{1.0-}\\\textbf{RAG}}  & \makecell{\textbf{Claude-}\\\textbf{Opus-}\\\textbf{RAG}} \\ 
\midrule
\multirow{3}{*}{\makecell[l]{Definition\\\ (F1 BoC $\uparrow$)}} & Def & \textbf{0.78} & \textbf{0.74} & \textbf{0.69} & \textbf{0.56} & \textbf{0.50} & \textbf{0.51} & \textbf{0.48} \\
                            & MultM & 0.47 & 0.42 & 0.47 & 0.39 & 0.37 & 0.48 & 0.40 \\ 
                            & NoM & 0.54 & 0.35 & 0.54 & 0.35 & 0.37 & 0.48 & 0.46 \\ 
\addlinespace
\multirow{3}{*}{\makecell[l]{Presence\\(ACC $\uparrow$)}} & Def & 0.67 &  0.58 & 0.5 & 0.50 & 0.42 & 0.50 & \textbf{0.50} \\
                            & MultM & 0.73 & \textbf{0.68} & \textbf{0.78} & \textbf{0.55} & \textbf{0.53} & \textbf{0.63} & \textbf{0.50} \\ 
                            & NoM & \textbf{0.80} & 0.50 & 0.70 & 0.50 & 0.40 & 0.30 & \textbf{0.50} \\
\bottomrule
\end{tabular}%
\end{table*}

\section{Discussion}
\label{sec:discussion}
In testing various state-of-the-art models (at the time of writing) on DesignQA, we developed questions about what factors impact models' scores on the benchmark. For example, to what extent is a model's score on DesignQA influenced by the RAG system used? How does the way in which a component is mentioned in the rule document affect a model's ability to answer questions about the component? Are models able to better answer questions about engineering drawings when dimensions are shown via scale-bars or when they are directly indicated on the drawing? Does additional image and textual context help models answer questions about rule compliance? We explore these questions in this section, and we propose general suggestions for how to develop models that are better suited to answering the questions presented within DesignQA.

\textbf{Impact of RAG System on Models' DesignQA Scores}
The focus of our work was to develop a benchmark to assess MLLMs' understandings of technical documentation and to utilize the benchmark to illustrate strengths and weaknesses of state of the art models, at the time of writing. As such, we did not invest time into tuning or developing RAG solutions designed specifically for DesignQA; development of useful RAG methods is a science within itself.~\cite{li2022survey} However, one of our findings in the previous section was that the simple LlamaIndex RAG used for the *-RAG models often did not provide the MLLM with useful context from the FSAE rule document.

To help disentangle RAG efficacy from inherent model performance, we generated GuaranteedRAG for each question in the benchmark. Created using a series of scripts with key-word matching, GuaranteedRAG is \textit{guaranteed} to contain context that is relevant to the question at hand. For example, in the case of the Retrieval questions, GuaranteedRAG context is designed to contain the rule-in-question. More details about the implementation of GuaranteedRAG can be found in Appendix B. Due to the cost of running the closed-source models, we only tested the open-source model, LLaVA-1.5, with GuaranteedRAG (LLaVA-1.5-GuaranteedRAG). We generated our GuaranteedRAG five different times (to account for randomness, see Appendix B), and we tested LLaVA-1.5 with each of these five GuaranteedRAGs, reporting averages and standard deviations across the five GuaranteedRAGs. 

The results of this experiment can be seen in Table~\ref{tab:idealrag_results}. While we expected LLaVA-1.5's score to improve substantially when tested with GuaranteedRAG instead of simple RAG, LLaVA-1.5-GuaranteedRAG -- across most subsets of the benchmark -- performs only slightly better than LLaVA-1.5-RAG. When compared with all the other models tested (Table 2), LLaVA-1.5-GuaranteedRAG does not perform the best across any subset of the benchmark. This finding underscores that much of the DesignQA score, at least in the case of LLaVA-1.5, is a result of inherent model capabilities (and lack thereof) rather than RAG efficacy.

The biggest improvement in score between LLaVA-1.5-RAG and LLaVA-1.5-GuaranteedRAG is for the Retrieval questions (0.112 versus 0.699 F1 BoW). Given useful rule document context, the Retrieval questions become straightforward. However, despite receiving the rule-in-question verbatim within an 8716-character text snippet (about 3\% of the total rule document), LLaVA-1.5-GuaranteedRAG still struggles to reliably return the rule-in-question for the Retrieval questions. LLaVA-1.5-GuranteedRAG still performs worse on the Retrieval questions than the best performer, GPT-4o-AllRules, which has to search for the relevant rule in the much longer, complete rule text. This finding illustrates LLaVA-1.5's inferior ability, compared with GPT-4o's, to extract relevant information from a body of text.

While receiving relevant context helps LLaVA-1.5 perform (on the whole) better on DesignQA, this experiment underscores that high quality RAG may not be the most important factor for improving MLLM scores on the benchmark. LLaVA-1.5 would need significant improvements in inherent model skills -- image analysis, engineering knowledge, extraction of relevant information from sections of text -- in order to improve its score further on DesignQA. We hope that GuaranteedRAG -- which helps isolate the testing of inherent model skills from the testing of RAG implementation -- can serve as a useful diagnostic tool for researchers seeking to improve foundation models.

\begin{table*}[th]
\centering
\caption{Effect of dimensioning system used in engineering drawings on model accuracy on Dimension QAs. These results highlight the fact that the MLLMs tested are generally better able to answer questions about engineering drawings that are direct-dimensioned rather than those that have a scale-bar. An exception is Gemini-1.0-RAG, which performs better on the scale-bar-dimensioned drawings.}
\label{tab:dimension_type}
\begin{tabular}{lccccccc}
\toprule
\multirow{0}{*}{\makecell{\textbf{ACC by} \\ \textbf{Dimension} \\ \textbf{System}}} & \multicolumn{7}{c}{\textbf{Model}} \\ 
\cmidrule(lr){2-8}
 & \makecell{\textbf{GPT-4o-}\\\textbf{AllRules}} & \makecell{\textbf{GPT-4-}\\\textbf{AllRules}} & \makecell{\textbf{GPT-4o-}\\\textbf{RAG}} & \makecell{\textbf{GPT-4-}\\\textbf{RAG}} & \makecell{\textbf{LLaVA-}\\\textbf{1.5-}\\\textbf{RAG}} &
\makecell{\textbf{Gemini-}\\\textbf{1.0-}\\\textbf{RAG}} & 
\makecell{\textbf{Claude-}\\\textbf{Opus-}\\\textbf{RAG}}\\
\midrule
Direct & \textbf{0.79} & \textbf{0.66} & 0.66 & \textbf{0.45} & \textbf{0.41} & 0.51 & \textbf{0.54} \\
Scale-bar & 0.75 & 0.28 & \textbf{0.70} & 0 & 0.40 & \textbf{0.55} & 0.45\\
\bottomrule
\end{tabular}
\end{table*}

\textbf{Effect of Vehicle Component Mention-Type on the Rule Comprehension Score}

We were curious to understand if the manner in which and the number of times a vehicle component is mentioned in the rule document impacts a model's ability to answer questions about that component. For the Rule Comprehension questions,
the component-in-question in each QA was tracked based on how it was mentioned in the rule document (definition-component, multi-mention component, or no-mention component), as explained previously in Section \ref{sec:comprehension}. 
Table \ref{tab:definition_mentions} breaks down each model's score for the Rule Comprehension questions (both Definition and Presence questions) by the three different component-mention types. 

There is a clear trend: for the Definition questions, where the model is asked to identify the name of a highlighted vehicle component in an image, each model always scores the highest on the definition components. Recall that for the Definition questions, the *-RAG models did not receive any context from the rule document, since the simple RAG implemented always returned the same section of the rule document (as the text of the prompt was identical for each Definition question). The models' higher scores on the definition-component questions within the Definition subset might then be explained by the fact that the models have seen similar images of highlighted definition-components during their pre-training. For example, images exist online that highlight the location of vehicle components like the main hoop and the front hoop. Since the definition-components are somewhat specific terms to the Formula SAE competition, they tend to be the focus of the online content and visuals that discuss the Formula SAE rules.

For the Presence questions -- where the model is asked to identify (yes/no) whether the specified component is present in an image -- the models exhibit a different trend. Almost all models score the highest on the multi-mention components (except for GPT-4o-AllRules and Claude-Opus-RAG, which scores equivalently on all component types). These questions show the models views of the vehicle -- zoomed views -- for which there are no similar images online. As such, the models are best able to visually identify the multi-mention components, which tend to be more common vehicle terms (like pedal assembly, brake system, and rocker arms).

Regardless of the specific reason for this observed pattern, the results presented in Table \ref{tab:definition_mentions} reflect that there are trends associated with component-mention type in the rule document and model score on the Rule Comprehension questions. It would be interesting to explore this trend further in future work.

\textbf{Effect of Engineering Drawing Dimensioning System on Models' DesignQA Scores}

As explained in Section \ref{sec:cad_rep}, one-third of the questions comprising the Dimension QAs used scale-bars in the engineering drawings, while the other two-thirds used direct dimensions. We were curious to see what impact these two different dimensioning systems had on model performance. As seen in Table \ref{tab:dimension_type}, most models perform better on the direct-dimensioned drawings than on the scale-bar drawings. The models seemed to struggle with the scale-bar. For example, GPT-4 sometimes indicated in its explanations that it was using the scale bar for ``estimated'' dimensions rather than precise ones and in some cases explained that its ``image capabilities...do not include measuring dimensions.'' 
GPT-4's difficulty with image measurement seems to have been improved in the newer version of the model, GPT-4o. GPT-4o-RAG actually had a higher accuracy score for the scale-bar-dimensioned images than for the direct-dimensioned images. 
Gemini-1.0-RAG was also an outlier, in that it performed better on the scale-bar dimensioned questions than on the direct dimensioned questions. Its ability to answer some of the scale-bar questions correctly likely contributed to the fact that it was the second best performer of the *-RAG models tested.  
Future research and model development must prioritize this feature, ensuring models can accurately interpret and apply dimensional data across various engineering tasks. This focus will enhance the precision and applicability of MLLMs in complex, real-world engineering environments.

\textbf{Effect of Additional Context on Models' Dimension Subset Scores}

We explained in Section \ref{sec:rule_compliance} that half of the Dimension questions were given additional context that we believed would help in answering the question while the other half were not. This additional context was in the form of highlights in the multiview CAD images -- on vehicle components pertaining to the rule-in-question -- and corresponding prompt text that revealed the name of the highlighted component (e.g., ``the front hoop is highlighted in pink''). Surprisingly, we did not see any obvious trends in model performance with versus without the additional context. GPT-4o-AllRules, GPT-4-AllRules, and LLaVA-1.5-RAG performed worse with additional context, while GPT-4o-AllRules, GPT-4-RAG, Gemini-1.0-RAG, and Claude-Opus-RAG performed better. More investigation into the effect of additional context is needed.

\textbf{Future Work: Recommendations for Models for This Benchmark}

In the results presented thus far, we have demonstrated how state-of-the-art models, at the time of writing, perform against our benchmark. Here, we present some observations on how one might modify these models to achieve improved performance on our benchmark. 

Firstly, inherent MLLM model skills must be improved in order to achieve higher scores on DesignQA. The models we tested with *-AllRules and *-GuaranteedRAG illustrate that even when given the relevant rule context, MLLMs have skills that need to be strengthened in order to improve their performances on design-according-to-technical-documentation-type tasks. We observed that the models tested struggle to reliably reference rules in the Formula SAE rule document. Rule extraction is a critical skill that models should be able to perform flawlessly; failure to do so could result in catastrophic consequences for subsequent rule compliance questions. Fine-tuning methods, like LoRA~\cite{hu2021lora}, QLoRA~\cite{dettmers2024qlora}, or InstructLAB~\cite{sudalairaj2024lab}, could be performed on models using the Retrieval QAs. This process may help MLLMs better understand and ``memorize'' the rule document, resulting in improved performances across the entire benchmark. 

Fine-tuning could also be used to improve MLLMs’ visual recognition of technical components and to strengthen their abilities to analyze engineering drawings. However, in our context, fine-tuning has several limitations. It requires a large amount of data that is not always available, typically struggles with tasks requiring precise recall, and becomes ineffective when design documents are frequently updated. Approaches like retrieval-augmented generation or leveraging expanded context windows of newer models align better with the needs of design-requirement tasks without the constraints associated with fine-tuning.

Second, improvement of the RAG model is critical so that relevant portions of the rule document can be accessed. We saw that GPT-4o-AllRules and GPT-4-AllRules, given all the context, almost always performed better than their respective -RAG models, which were given the selected rule document pages through simple retrieval. While one might argue that these findings illustrate that models with longer context lengths should be developed and used, it is important to note that using models with longer context inputs is significantly more expensive, as of the time of writing. The cost to feed the entire rule document into the context for each question was more than ten times the cost of selecting relevant pages via RAG. As such, effective RAG could help reduce the computational burden and provide useful context to the user regarding relevant sections in the original documents. 

While MLLMs coupled with GuarnateedRAG showed improvements over those paired with simple RAG, the GuarnateedRAG solution -- which was custom-developed for the questions in DesignQA using key-word matching scripts -- is not generalizable to other kinds of design-requirement-type questions. Multimodal RAG might be an interesting avenue for exploration, since the images in the VQAs could help in selecting the relevant portion of the rule document. The RAG could also perhaps be improved by experimenting with different chunking methods. Nevertheless, as models become cheaper, more computationally efficient, and trained with larger context windows, RAG approaches might become less useful.

Engineering design, inherently grounded in practical applications, necessitates a deep understanding of technical documents to ensure that designs not only meet but also adhere to stringent requirements and standards. One of the critical insights emerging from our benchmark evaluations is the realization that modern AI models are still in the early stages of truly understanding engineering documentation. This limitation highlights a significant gap in the path toward fully automated design processes. There is therefore substantial future work to be done to improve MLLM models' abilities to solve engineering design and design requirement problems.

\textbf{Limitations}

While this benchmark presents a first step into formally evaluating a model's ability to interpret and understand design requirement documentation, DesignQA does focus on just a single rule document, and it is unknown how the results on DesignQA would vary if a different technical document was used for dataset curation. Unfortunately, benchmark development costs (requiring expert manual QA creation and validation) pose limitations on scaling the size of the dataset or the number of technical documents referenced. Our benchmark is also limited to six types of design-requirement-based tasks, while there are certainly others that are worth exploring. However, we believe that the six subsets of our benchmark are reflective of the primary ways in which engineers consult design requirement documentation. We also inherently made choices about prompt format and CAD/engineering-drawing image presentation in creating our benchmark. DesignQA results may not generalize to similar questions where the prompts or images are tweaked. Exploring the effect of other prompt and image formats on models' scores on DesignQA could be the subject of future studies. Overall, generalizability (to other similar questions not contained within the test dataset) is a common limitation of LLM benchmarks~\cite{mcintosh2024inadequacies}. While generalizability may be a limitation, DesignQA is an attempt to track MLLMs' abilities to answer questions about design requirements.

In terms of the validity of DesignQA, the automatic evaluation metrics we used could be viewed as a limitation. There are cases where a model response's score on an automatic evaluation metric may not match the validity a human would assign to the same response. For example, instances where a model doesn't follow the specified response format (e.g. not separating Compilation question rules with commas) result in artificially low scores for the model on that metric. There were also cases, for the Retrieval questions, where models returned rules that included child rules (which is not technically wrong), whereas our ground truth reference did not include child rules. Evaluating semantic equivalence is a difficult problem, and the BLEU and ROUGE scores used to evaluate model explanations have known limitations. Collecting more human-generated reference explanations, while time consuming, would allow for more accurate scoring of model explanations. While automated evaluation metrics may slightly artificially penalize models, the alternative -- manually scoring responses -- would be extremely resource-intensive for thousands of questions and may also have many biases. In fact, automatic evaluation is necessary for a benchmark dataset that is meant to be used by the community to continuously test models developed in the future.

Another limitation of DesignQA that may affect benchmark validity is the small size of certain benchmark subsets. While there are more than a thousand Retrieval questions -- since they could be generated automatically -- the other subsets had to be manually generated and thoroughly reviewed and are therefore smaller in size. In particular, the Functional Performance questions only have 16 QAs due to the difficulty of obtaining vehicle testing data that was immediately relevant to the requirement document. While some of the benchmark subsets may be small, we prioritized manual, high-quality expert QA creation and validation over alternative methods -- like synthetic data generation or crowdsourcing -- which may compromise dataset quality.

\section{CONCLUSION}
This study introduces \acr{}, a novel MLLM benchmark with 1451 questions and answers based on data from MIT Motorsports and the FSAE competition rule document. The benchmark is designed to assess large language models' abilities to answer questions about design according to technical requirements. The benchmark is divided into six subsets - Retrieval, Compilation, Definition, Presence, Dimension, and Functional Performance - representative of tasks performed by engineers when designing according to technical specifications. Each subset of the benchmark has its own automatic evaluation metric, so that new MLLMs can be seamlessly tested. The questions in the benchmark are designed by humans - MIT Motorsports members, industry professionals, and researchers - to ensure a high-quality benchmark. Different from many existing MLLM benchmarks, \acr{} contains document-grounded VQAs, in which the input image and document come from differing sources, characteristic of many real-world scenarios. 

Using \acr{}, we conducted a rigorous evaluation on variants of several state-of-the-art (at the time of writing) MLLMs: GPT-4o, GPT-4, Claude-Opus, Gemini-1.0, and LLaVA-1.5. While GPT-4o was the best performing model of the MLLMs tested, we uncovered significant limitations in MLLMs' current abilities to accurately interpret complex technical documents, specifically in referencing relevant requirements and in  analyzing technical images. These findings illustrates that analyzing and designing according to technical documentation requires a diverse skill-set, and that there is a need to develop MLLMs better suited to the multifaceted-nature of engineering design. Our research highlights the need for advancements in AI models to enhance their comprehension of engineering requirements and documentation, suggesting directions for future efforts. Our work aims to bridge the gap in AI's capability to support engineering design processes more effectively, paving the way for sophisticated AI-assisted engineering solutions.

\bibliographystyle{asmeconf}

\section*{Acknowledgments}
The authors wish to express their gratitude to Henry Smith and the MIT Motorsports team for their critical contribution to dataset creation and verification, which significantly enhanced the quality of this research. This material is based upon work supported by the National Science Foundation Graduate Research Fellowship under Grant No. 2023345746. Any opinion, findings, and conclusions or recommendations expressed in this material are those of the authors(s) and do not necessarily reflect the views of the National Science Foundation.

\begin{nomenclature}
\entry{Claude-Opus}{Refers specifically to \textit{claude-3-opus-20240229}}
\entry{Gemini-1.0}{Refers specifically to \textit{models/gemini-1.0-pro-vision}}
\entry{GPT-4}{Refers specifically to \textit{gpt-4-1106-vision-preview}}
\entry{GPT-4o}{Refers specifically to \textit{gpt-4o}}
\entry{FSAE}{Formula SAE}
\entry{LLaVA-1.5}{Refers specifically to \textit{llava-1.5-13b}}
\entry{LLM}{Large Language Model}
\entry{MLLM}{Multimodal Large Language Model}
\entry{RAG}{Retrieval-Augmented Generation}
\entry{-RAG}{Models tested with LlamaIndex's simple RAG framework}
\entry{VQA}{Visual Question Answer (Benchmark)}
\end{nomenclature}

\bibliography{references}

\section*{Appendix A: Model Parameters}
All MLLMs evaluated by the authors were run through the LlamaIndex framework. We used the default parameters, set in LlamaIndex, for each model tested. For LLaVA-1.5, the default parameters are temperature of 0.75 and top\_p of 0.9.\footnote{https://github.com/run-llama/llama\_index/blob/main/llama-index-integrations/multi\_modal\_llms/llama-index-multi-modal-llms-replicate/llama\_index/multi\_modal\_llms/replicate/base.py} For GPT-4o, GPT-4, Gemini-1.0, and Claude-Opus, the default temperature set by the LlamaIndex framework is 0.1.\footnote{https://github.com/run-llama/llama\_index/blob/main/llama-index-core/llama\_index/core/constants.py} For these models, LlamaIndex doesn't set defaults for top\_p, so each model should retain its own default top\_p value. For the Rule Extraction questions, max\_new\_tokens was set to 250 for all models. For the Rule Comprehension questions, max\_new\_tokens was set to 100. For the Rule Compliance questions, max\_new\_tokens was set to 1500.

\section*{Appendix B: GuaranteedRAG}

\textbf{GuaranteedRAG for Retrieval Questions:} For each Retrieval question, an 8716-character \footnote{Note that in generating the simple LlamaIndex RAG for most of the questions, except the Rule Compliance questions, the top 15 most relevant document chunks were used. These top 15 most relevant chunks had an average character count of 10895 characters. Initially, we tried generating 10895 character count excerpts for GuaranteedRAG, but these contexts, coupled with the prompts, often exceeded the token limit for LLaVA. This happened due to diffrences in character counts and token counts. So as to not run up against token limits, we reduced the GuaranteedRAG context length to 8716 characters for all questions. 8716 characters corresponds with the average length of the top 12 most relevant document chunks in the simple LlamaIndex RAG implementation.} portion of the text was extracted from the full rulebook. This portion was specifically selected to contain the rule-in-question; the location of the rule-in-question within the excerpt was randomized.

\textbf{GuaranteedRAG for Compilation Questions:} The ground truth answer to each Compilation question is a list of rule numbers. For each Compilation question, we extracted the text for each of these ground truth rules from the rule document. If the total sum of these rule texts was less than 8716 characters, we randomly added in other rules until we reached the 8716 limit. If, on the other hand, the total sum of the ground truth rule texts was greater than 8716 before adding additional, random rules, we iteratively removed the longest rule from the list until we met the character limit. The rule texts were randomly shuffled (before appending them together) so that they were presented in no specific order. 

\textbf{GuaranteedRAG for Definition and Presence Questions:} As explained in Section \ref{sec:comprehension}, each Definition and Presence question tests the model's understanding of a vehicle component. The components are either: 1) mentioned explicitly in the definition section of the rule document (``definition component''), 2) mentioned multiple times throughout the document but are not explicitly in the definition section (``multi-mention component''), or 3) are not mentioned in the rule document at all (``no-mention component''). For each Definition or Presence question, we generate GuaranteedRAG in accordance with the component's mention type. For definition component questions, we extract the two ``Definition'' sections of the rule document and randomly place them in between two other randomly selected segments of rule text, such that the total extracted text length is 8716 characters. For multi-mention component questions, we find every mention of the component within the rule document. For each mention, we extract a constant length text excerpt from the rule document with the mention centered in the excerpt. When appended together, the total length of these text excerpts is 8716 characters. For no-mention component questions, we randomly select an 8716-character segment from the rule document.

\textbf{GuaranteedRAG for Dimension and Functional Performance Questions:} Each of these questions asks the model whether a design conforms with a specific rule number. For each of these questions, an 8716-character portion of the text was extracted from the rulebook. This portion is guaranteed to contain the rule text of the rule in question; the location of the rule-in-question within the excerpt was randomized. In some cases, the rule in question references another rule in the FSAE rule document. In these cases, the GuaranteedRAG is split into two even chunks that are appended together to form the full 8716-character GuaranteedRAG: the first half is guaranteed to contain the rule text for the rule (at a random location) and the second half is guaranteed to contain the rule text for the referenced rule (at a random location).

\end{document}